# A Machine-Learning Surrogate for Component Criticality Ranking in Interdependent Power–Communication Networks

Sohini Roy
*Department of Computer Science*
*University of Nevada, Las Vegas*
Las Vegas, USA
sohini.roy@unlv.edu

Xheni Hylviu
*Department of Computer Science*
*University of Nevada, Las Vegas*
Las Vegas, USA
hylvix1@unlv.nevada.edu

***Abstract*—Cyber-physical power systems are vulnerable to cascading failures caused by interdependencies between power and communication infrastructures. Because evaluating large $N-k$ contingency sets with a high-fidelity simulator is computationally expensive, this paper develops a machine-learning surrogate using the previously published Modified Implicative Interdependency Model (MIIM) as the ground-truth cascade simulator. The surrogate predicts contingency severity from leakage-free structural features and derives an association-based component-criticality ranking for resilience screening. On the IEEE 118-bus system, Gradient Boosting achieves a held-out Spearman correlation of 0.849 for contingency-severity ranking. Using five-fold out-of-fold predictions, the resulting component ranking achieves a Spearman correlation of 0.838 with the MIIM-derived ranking and closely approaches the observed cross-sample reproducibility level. Feature-ablation results show that inter-layer dependency features drive most of the surrogate's advantage, while end-to-end screening is approximately 158× faster than direct MIIM evaluation. The results support a two-stage workflow in which the surrogate screens candidate contingencies and components, and MIIM provides selective verification rather than directly identifying optimal hardening actions.***

## I. Introduction

Cyber-physical power systems (CPPS) depend on close coordination between the physical grid and communication infrastructure for monitoring, control, and data exchange. Although this coupling improves situational awareness and operational efficiency, it also creates cascading-failure pathways: power-layer disruptions can disable communication entities, while communication failures can degrade grid observability and control. Resilience analysis must therefore capture interdependent failure propagation rather than treating the two layers separately.

A central challenge is evaluating $N-k$ contingencies, in which multiple components fail simultaneously and trigger cross-layer cascades. High-fidelity interdependency simulation can assess individual contingencies, but exhaustive evaluation becomes computationally prohibitive as network size and contingency order increase. Resilience planning therefore requires efficient methods to identify severe contingencies and screen components associated with severe cascade outcomes for subsequent intervention-based analysis.

This paper addresses that need using the previously published Modified Implicative Interdependency Model (MIIM) [1] as the ground-truth simulation engine rather than as a new contribution. MIIM represents intra- and interdependencies in joint power–communication systems and evaluates cascade states under failures. Using MIIM-generated severities as labels, we train a machine-learning surrogate on leakage-free structural features and derive an association-based component-criticality ranking for resilience screening.

Three findings shape the contribution. First, on the IEEE 118-bus system, the surrogate achieves a held-out Spearman correlation of 0.849 for contingency-severity ranking. Using five-fold out-of-fold predictions, the component ranking attains 0.838 against the MIIM-derived ranking, closely approaching the mean cross-sample MIIM reproducibility of 0.854 without using in-sample predictions. Second, full-topology baselines provide meaningful but lower global agreement, with correlations of 0.599-0.689, while feature ablation shows that the surrogate's advantage is driven primarily by inter-layer dependency information. Third, end-to-end screening, including feature extraction and inference, is approximately 158 × faster than direct MIIM evaluation. These results support a two-stage workflow in which the surrogate rapidly screens contingencies and components, while MIIM is retained for selective verification of intervention and hardening decisions.

The main contributions of this paper are as follows:

*1) A fast, leakage-free machine-learning surrogate for MIIM-based $N-k$ contingency-severity prediction in interdependent power–communication networks.*

*2) An association-based component-criticality ranking computed from five-fold out-of-fold predictions, ensuring that no in-sample predictions are used in the reported component-level evaluation.*

*3) A cross-sample reproducibility analysis showing that the surrogate closely approaches the observed MIIM reproducibility reference.*

*4) A comparison with full-topology baselines, an ablation-based assessment of inter-layer dependency features, and an end-to-end speedup evaluation relative to direct MIIM simulation.*

## II. Background and Related Work

Power system resilience has been widely studied in response to extreme weather events, physical disruptions, and cyber threats affecting modern grids. Bie et al. [2], Gholami et al. [3], and Bhusal et al. [4] clarify that resilience extends beyond traditional reliability and security by emphasizing the ability of the grid to withstand, adapt to, and recover from high-impact disturbances. Ashok et al. [5] further show that modern grid resilience must be treated from a cyber-physical perspective because monitoring, protection, control, and communication infrastructures materially affect operational continuity under adverse conditions.

A related line of work studies the grid as a complex or cyber-physical network and analyzes resilience using structural or graph-theoretic measures of component importance. Moradi Amani and Jalili [6], for example, review resilience and reliability analysis of power grids from a complex-networks perspective and discuss the use of centrality-type measures for identifying influential components. Such approaches are attractive because they are computationally inexpensive and structurally interpretable, but in interdependent power-communication systems an important empirical question is how much of the true cascade-driven criticality ordering these topology-based measures actually recover—a question we address quantitatively in Section V. Machine-learning methods have also been used for online security assessment and contingency screening. Surveys by Mehrzad et al. [7] and De Caro et al. [8] show the growing role of learned surrogates in approximating expensive simulation-based analyses, while Sunitha et al. [9] demonstrated early use of neural networks for online static security assessment and severity ranking. More recently, Zheng et al. [10] proposed a physics-informed GNN for AC power-flow-based $N-2$ screening, but that work targets the power system in isolation and focuses on feasibility classification rather than component criticality in a coupled power-communication setting.

The present work builds most directly on prior MIIM-based literature. Roy et al. [1] introduced the Modified Implicative Interdependency Model (MIIM) to represent intra- and interdependencies in joint power-communication networks using multi-valued interdependency relations. Roy and Sen later used MIIM for identifying K-most vulnerable entities [11] and for constructing a self-updating K-contingency list [12]. These studies established MIIM as a useful framework for interdependency-aware critical-entity analysis, but they relied on direct MIIM evaluation rather than a scalable learned surrogate.

In contrast, this paper uses the previously published MIIM only as the ground-truth simulation engine and develops a learned surrogate for fast $N-k$ contingency-severity prediction in interdependent power–communication networks. The surrogate is then used to derive an association-based component-criticality ranking and is evaluated against both the MIIM-derived reference ranking and topology-based baselines computed on the full interdependent network. More importantly, the paper quantifies the cross-sample reproducibility of MIIM-derived component criticality under independent contingency sampling and uses the observed mean agreement as an empirical reproducibility reference. This reframes the comparison from whether the surrogate merely outperforms topology to how closely it approaches the component-ranking agreement observed across independently sampled contingency datasets.

## III. Methodology

This work builds on the previously published Modified Implicative Interdependency Model (MIIM) [1], which is used here only as the ground-truth simulation engine for contingency evaluation. MIIM represents a joint power–communication system through multi-state interdependency relations and generates contingency-severity labels. These labels are approximated by a learned surrogate using pre-simulation structural features for rapid contingency screening and association-based component-criticality ranking.

### A. MIIM-Based Label Generation

Contingency severity is defined as:

$$S(F) = 1 - \frac{1}{T}\sum_{t=1}^{T} O_t \qquad (1)$$

where $F$ is the initiating $N-k$ contingency, $T = 10$ is the cascade horizon, and $O_t \in [0,1]$ is the mean operability across all in-scope nodes at cascade time step $t$, with failed nodes contributing zero operability. The immediate post-injection state $O_0$, which reflects the initiating failures before cascade propagation, is excluded from the averaging horizon. Thus, $S(F) = 0$ denotes no sustained operability loss over the horizon, whereas $S(F) = 1$ denotes complete loss. Labels are generated using exhaustively enumerated low-order contingencies and stratified samples of higher-order contingencies.

### B. Leakage-Free Structural Feature Design

Each contingency is represented by thirteen pre-simulation features derived solely from the initiating failed set and network structure: contingency order $k$; counts of failed power, communication, and other nodes; the number and fraction of severed inter-layer edges; gateway involvement and the failed-gateway fraction; maximum failed-node degree and failed-subgraph connectivity; failed-power and failed-communication ratios; and an encoded event type. Here, $k$ denotes the number of simultaneously failed components in the $N-k$ contingency. To prevent target leakage, all post-cascade quantities are excluded, including operability traces $O_0$–$O_{10}$, post-cascade failure counts, stabilization statistics, and the severity score and class labels.

### C. Surrogate Model and Evaluation

Linear Regression, Random Forest, and Gradient Boosting are evaluated using an 80/20 severity-class-stratified split. Severity classes–none, mild, moderate, and severe; are obtained by applying 51-node reference cutpoints of 0.02, 0.10, and 0.25, scaled by $51/n$ for a network with $n$ in-scope nodes. The models predict continuous severity $S(F)$; severity class is used only for stratification. The primary metrics are Spearman rank correlation and contingency-level Precision@5%:

$$Precision@5\% = {|\mathcal{A}_{5\%} \cap \hat{\mathcal{A}}_{5\%}|}\Big/{|\hat{\mathcal{A}}_{5\%}|} \qquad (2)$$

where $\mathcal{A}_{5\%}$ and $\hat{\mathcal{A}}_{5\%}$ are the top 5% of contingencies under MIIM and surrogate-predicted severity, respectively. RMSE and $R^2$ are secondary metrics. Gradient Boosting performs best and is used for subsequent criticality and ablation analyses.

The 80/20 split is used for within-seed evaluation and feature ablation. Component-level evaluation follows the five-fold out-of-fold protocol in Section III-D, while cross-seed robustness is evaluated in Section V-B.

To evaluate the importance of inter-layer information, we remove the four inter-layer dependency features defined in Section III-B and retrain the model using the same data split and settings. The decrease in Spearman correlation relative to the full-feature model measures the combined contribution of these features.

### D. Component Criticality and Ranking Evaluation

For each interdependent network component $i$, MIIM-derived criticality is defined as:

$$C(i) = \frac{1}{|\mathcal{F}_i|} \sum_{F \in \mathcal{F}_i} S(F) \tag{3}$$

where $\mathcal{F}_i$ is the set of sampled contingencies whose failed set includes component $i$, and $|\mathcal{F}_i|$ is its cardinality. Only components with $|\mathcal{F}_i| \geq 5$ are scored. To avoid optimistic evaluation while retaining broad coverage, surrogate criticality is computed using five-fold severity-class-stratified out-of-fold predictions. In each fold, Gradient Boosting is trained on four folds and predicts the fifth, so every contingency is predicted by a model that did not train on it:

$$\hat{C}(i) = \frac{1}{|\mathcal{F}_i|} \sum_{F \in \mathcal{F}_i} \hat{S}_{OOF}(F) \tag{4}$$

where $\hat{S}_{\mathrm{OOF}}(F)$ uses only pre-simulation features. No in-sample predictions are used in the reported component rankings.

Both measures are inclusion-based averages and therefore represent association-based screening signals, not marginal or causal estimates of hardening benefit. A highly ranked component may co-occur with other damaging components; thus, the ranking identifies candidates for intervention-based MIIM verification rather than optimal components to harden. The surrogate is compared with three full-topology baselines: node degree, betweenness centrality, and degree with a gateway bonus of 5. Agreement with the MIIM-derived ranking is measured using Spearman correlation and Top-$K$ overlap for $K \in \{10, 20, 30\}$:

$$\mathrm{Overlap@}K = \frac{|\mathcal{T}_K \cap \mathcal{P}_K|}{K} \tag{5}$$

where $\mathcal{T}_K$ and $\mathcal{P}_K$ are the MIIM-derived and comparison top-$K$ sets, respectively.

## IV. Experimental Setup

The framework is evaluated on the IEEE 14-bus and IEEE 118-bus systems, containing 51 and 446 in-scope nodes, respectively, using the previously published MIIM benchmark communication overlay and interdependency relations. Contingencies up to $k = 4$ are considered. All $N - 1$ contingencies are enumerated, while higher-order contingencies are stratified-sampled.

For the 118-bus system, the main seed-42 dataset contains 10,446 contingencies: 446 $N-1$, 5,000 $N-2$, 3,000 $N-3$, and 2,000 $N-4$ cases. Seeds 43 and 44 each contain 3,946 contingencies, with corresponding per-order counts of 446, 2,000, 1,000, and 500, and are used for within-seed robustness, fixed-topology cross-seed robustness, and reproducibility analysis. For the 14-bus system, three datasets of 2,126 contingencies each are generated using the same framework. Severity classes follow Section III-C; for the 118-bus system, the scale $51/446 \approx 0.114$ yields cutpoints of approximately 0.0023, 0.0114, and 0.0285 on $S(F)$.

In the main 118-bus dataset, mean severity increases from 0.0068 at $k = 1$ to 0.0134, 0.0208, and 0.0275 for $k = 2, 3,$ and 4, respectively. This monotonic trend supports $k$ as a pre-contingency feature while retaining substantial within-order variability.

All experiments use scikit-learn 1.3.2. Gradient Boosting uses 100 estimators, maximum depth 5, and random state 42; Random Forest uses 100 estimators, maximum depth 10, and random state 42.

## V. Results and Discussion

### A. Surrogate Prediction Accuracy

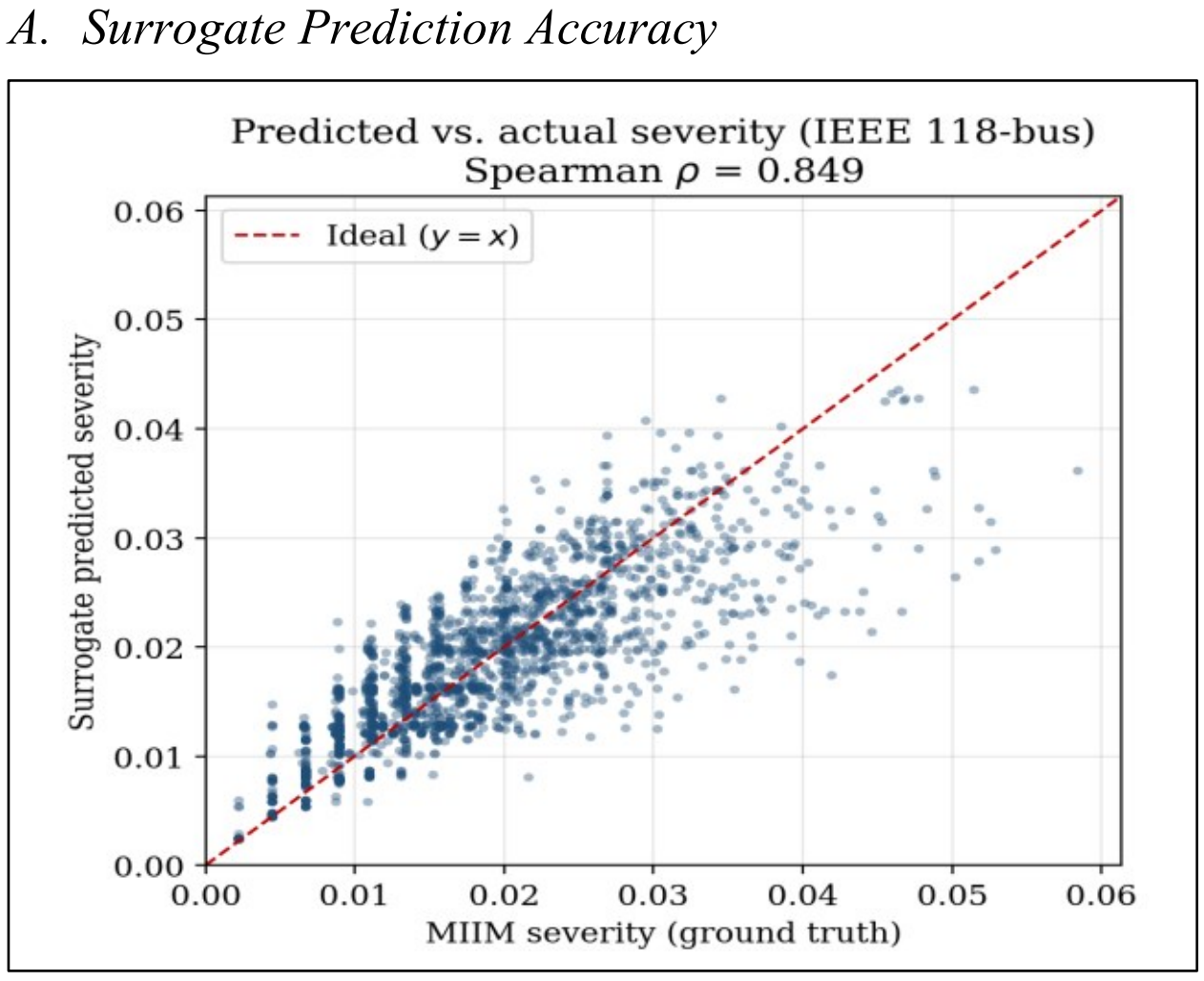

Fig. 1. Predicted versus actual contingency severity on the IEEE 118-bus system using the Gradient Boosting surrogate. The dashed red line denotes the ideal $y = x$ relationship.

Table I compares Linear Regression, Random Forest, and Gradient Boosting on the IEEE 14-bus and IEEE 118-bus systems. Gradient Boosting provides the strongest overall ranking performance on both networks. On the 118-bus system, it achieves Spearman = 0.849, Precision@5% = 36.5%, and $R^2 = 0.689$. Its RMSE of 0.0050 is essentially tied with Random Forest at 0.0051 and is therefore best interpreted as comparable to the best result rather than decisively lower. On the 14-bus system, Gradient Boosting achieves Spearman = 0.900, confirming its effectiveness for contingency-severity ranking.

Fig. 1 shows predicted versus actual severity on the 118-bus test set. Despite the compressed severity range, the predictions preserve the severity ordering, consistent with the Spearman correlation of 0.849 and the surrogate's intended screening role.

TABLE I. MODEL COMPARISON ON THE IEEE 14-BUS AND IEEE 118-BUS BENCHMARK SYSTEMS.

| Network | Model | RMSE | $R^2$ | Spearman | Precision@5% |
|---|---|---|---|---|---|
| 14-bus | Linear Regression | 0.0358 | 0.578 | 0.777 | 38.1% |
| 14-bus | Random Forest | 0.0264 | 0.772 | 0.894 | 47.6% |
| 14-bus | **Gradient Boosting** | **0.0257** | **0.784** | **0.900** | **42.9%** |
| 118-bus | Linear Regression | 0.0063 | 0.502 | 0.704 | 21.2% |
| 118-bus | Random Forest | 0.0051 | 0.678 | 0.843 | 33.7% |
| 118-bus | **Gradient Boosting** | **0.0050** | **0.689** | **0.849** | **36.5%** |

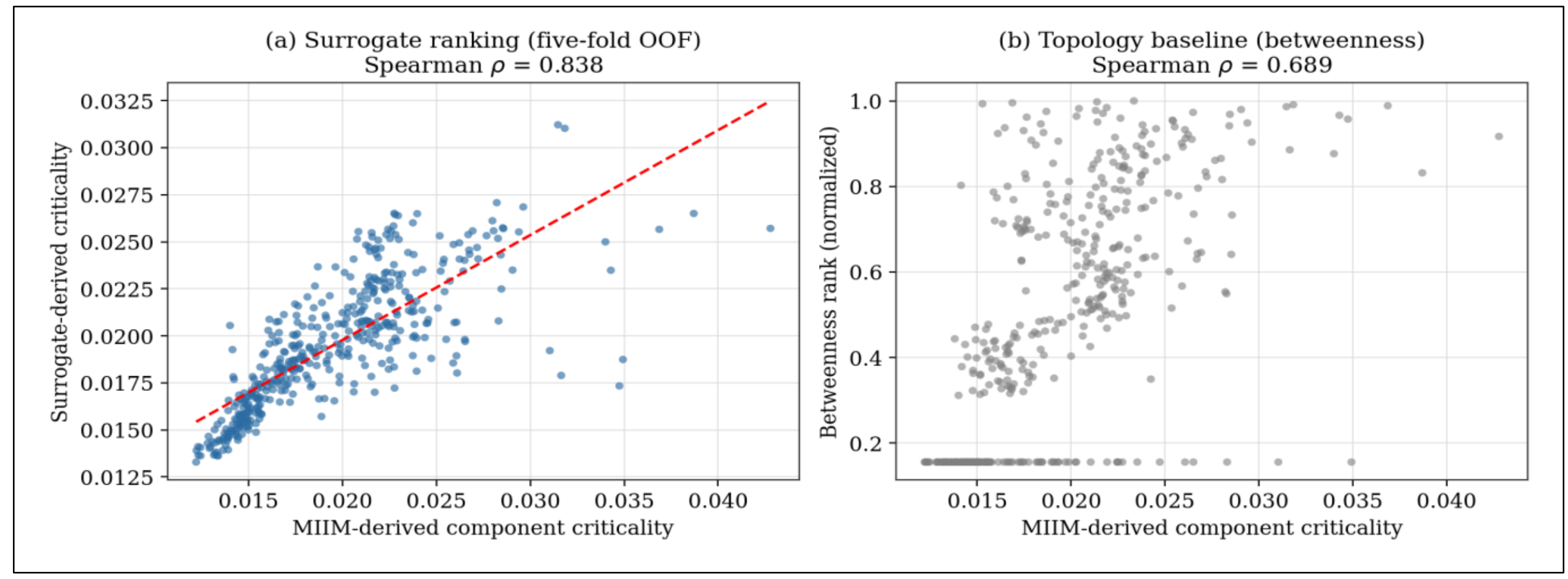

Fig. 2. Component-criticality ranking on the IEEE 118-bus system. The surrogate ranking, computed from five-fold out-of-fold predictions, closely agrees with the MIIM-derived ranking ($\rho$ = 0.838), while betweenness centrality, the strongest topology baseline, shows weaker global agreement ($\rho$ = 0.689).

## *B. Robustness Across Contingency Samples*

TABLE II. ROBUSTNESS ACROSS CONTINGENCY SAMPLES FOR THE GRADIENT BOOSTING SURROGATE

A. WITHIN-SEED HELD-OUT EVALUATION WITH RETRAINING

| Network | Seed-42 | Seed-43 | Seed-44 | Spread |
|---|---|---|---|---|
| 14-bus Spearman | 0.900 | 0.889 | 0.903 | 0.014 |
| 14-bus Precision@5% | 42.9% | 71.4% | 61.9% | – |
| 118-bus Spearman | 0.849 | 0.878 | 0.865 | 0.029 |
| 118-bus Precision@5% | 36.5% | 46.2% | 25.6% | – |

B. IEEE 118-BUS FIXED-TOPOLOGY CROSS-SEED EVALUATION WITHOUT RETRAINING

| Train →Test | Severity Spearman | RMSE | Precision @5% | Criticality Spearman |
|---|---|---|---|---|
| Seed 42 → Seed 43 | 0.866 | 0.0047 | 46.2% | 0.827 |
| Seed 42 →Seed 44 | 0.864 | 0.0047 | 40.6% | 0.827 |

*Precision@5% spread is omitted because this extreme-tail metric is sensitive to small changes in top-ranked membership.

Table II evaluates robustness under two protocols. In the within-seed analysis, Gradient Boosting is retrained and evaluated separately on each seed dataset using the same feature set, model configuration, and 80/20 stratified split. In the cross-seed analysis, the model is trained once on seed 42 and applied to seeds 43 and 44 without retraining.

Within-seed Spearman correlations range from 0.889 to 0.903 on the IEEE 14-bus system and from 0.849 to 0.878 on the IEEE 118-bus system, indicating stable ranking performance across independently sampled datasets. Precision@5% is more variable, as expected for an extreme-tail metric. Under the stricter cross-seed protocol, the seed-42 model retains severity-ranking correlations of 0.866 and 0.864 on seeds 43 and 44, respectively, with component-ranking correlations of 0.827 on both datasets. These results show that the learned relationships remain stable across independently sampled contingency sets without retraining. This analysis differs from Section V-C, which evaluates the cross-sample reproducibility of the MIIM-derived component ranking itself.

## *C. Component Criticality and Cross-Sample Reproducibility*

The central contribution of this paper is the use of the surrogate to derive a component-criticality ranking. Because MIIM-derived criticality is estimated from finite contingency sets, it is sample-dependent. To quantify this variation, criticality is computed independently on the three seed datasets, and agreement among the resulting rankings is measured using pairwise Spearman correlations. Unlike the robustness analysis in Section V-B, which retrains and evaluates the surrogate on each seed dataset, this analysis evaluates the reproducibility of the MIIM-derived ranking itself.

The pairwise Spearman correlations are 0.838, 0.860, and 0.863, yielding a mean cross-sample agreement of 0.854. This value is treated as an empirical cross-sample reproducibility reference under the present sampling pipeline rather than as a theoretical upper bound.

Using five-fold out-of-fold predictions, the surrogate attains Spearman = 0.838, closely approaching the mean cross-sample MIIM agreement of 0.854 without using in-sample predictions. Across five random fold partitions, the out-of-fold correlation ranges from 0.837 to 0.839, with a mean of 0.838, indicating negligible sensitivity to fold assignment. The full-topology baselines provide lower global agreement, with correlations of 0.689 for betweenness centrality, 0.648 for node degree, and 0.599 for degree with a gateway bonus of 5. Fig. 2 contrasts the surrogate ranking with betweenness, the strongest topology baseline.

TABLE III. COMPONENT-CRITICALITY RANKING ON THE IEEE 118-BUS SYSTEM.

| Method | Spearman | Top-10 | Top-20 | Top-30 |
|---|---|---|---|---|
| **Surrogate, five-fold OOF** | **0.838** | 30% | 40% | 53% |
| Baseline: betweenness | 0.689 | 30% | 35% | 30% |
| Baseline: node degree | 0.648 | 30% | 35% | 30% |
| Baseline: degree + gateway | 0.599 | 50% | 40% | 60% |

Betweenness is a meaningful topology baseline because components lying on many network paths may facilitate cascade propagation. However, the surrogate achieves stronger global agreement by combining structural and contingency-specific inter-layer information. The Top- $K$ results are more mixed, and no method dominates every cutoff. The degree-plus-gateway baseline performs better at Top-10 and Top-30 and matches the surrogate at Top-20, partly because six of the MIIM top-10 components are gateways and the bonus promotes all gateways. The current surrogate also misses some components that are inconspicuous under the structural feature representation. For example, PMU12 has degree zero in the graph used for feature extraction, yet ranks fourth under the MIIM-derived ranking and 247th under the out-of-fold surrogate ranking. These limitations motivate richer dependency-aware network representations; until then, the surrogate is best used for broad screening, with MIIM retained for final intervention-based verification.

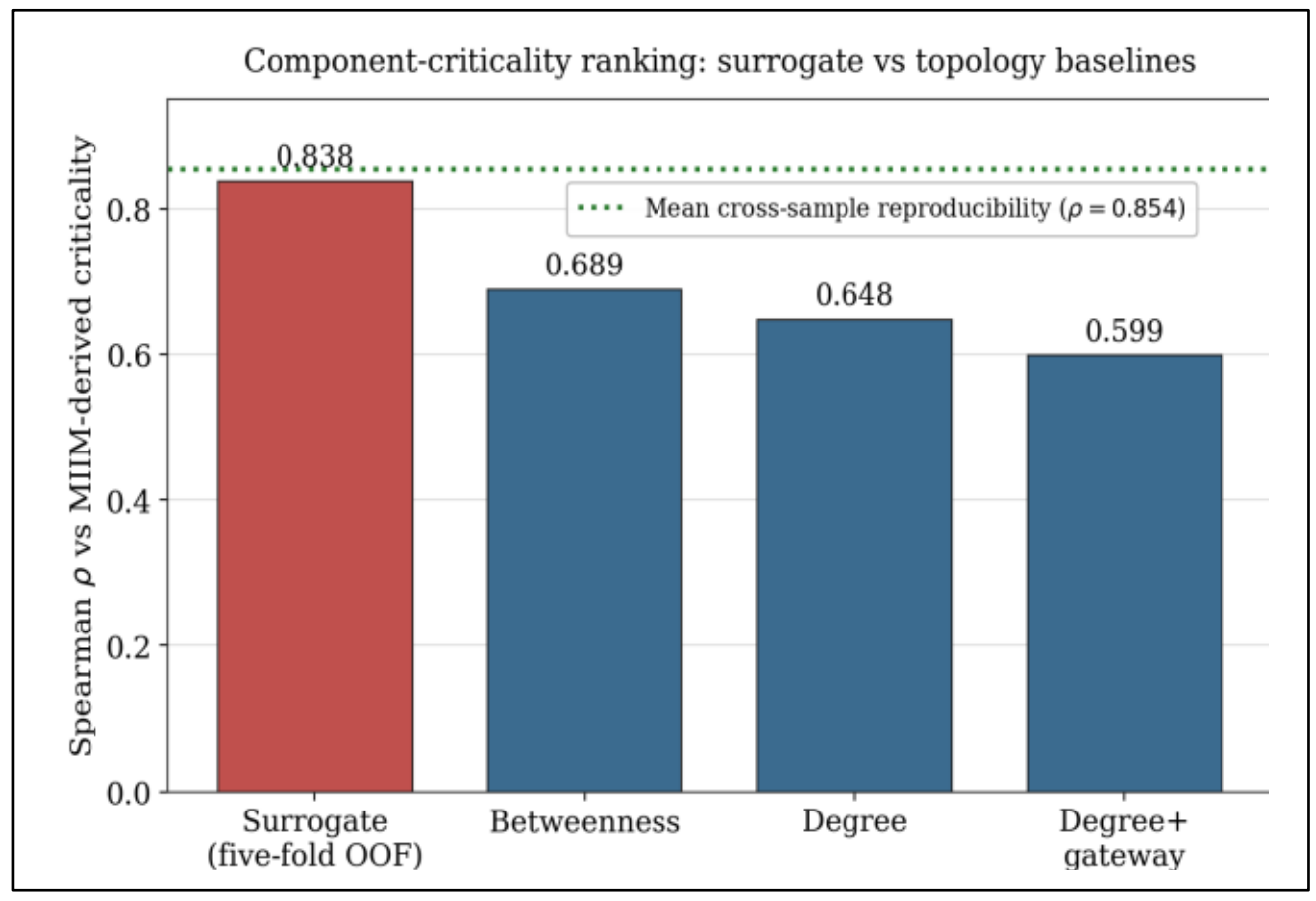

Fig. 3. Spearman correlation of the surrogate and topology-based component-criticality rankings with the MIIM-derived reference ranking on the IEEE 118-bus system. The dotted line marks the mean cross-sample MIIM agreement ($\rho = 0.854$).

## D. What the Surrogate Learns

Fig. 4 shows the Gradient Boosting feature importances. The fraction and number of severed inter-layer edges are the most influential features, followed by contingency order $k$, indicating that cross-layer disruption provides information beyond raw network connectivity.

Table IV confirms this result. Removing the four inter-layer features reduces Spearman from 0.849 to 0.799, whereas removing the raw-topology features reduces it to 0.813. Inter-layer features alone still achieve Spearman $= 0.763$, compared with 0.657 using only $k$. Thus, both feature groups contribute, but the surrogate's performance is driven primarily by inter-layer dependency information.

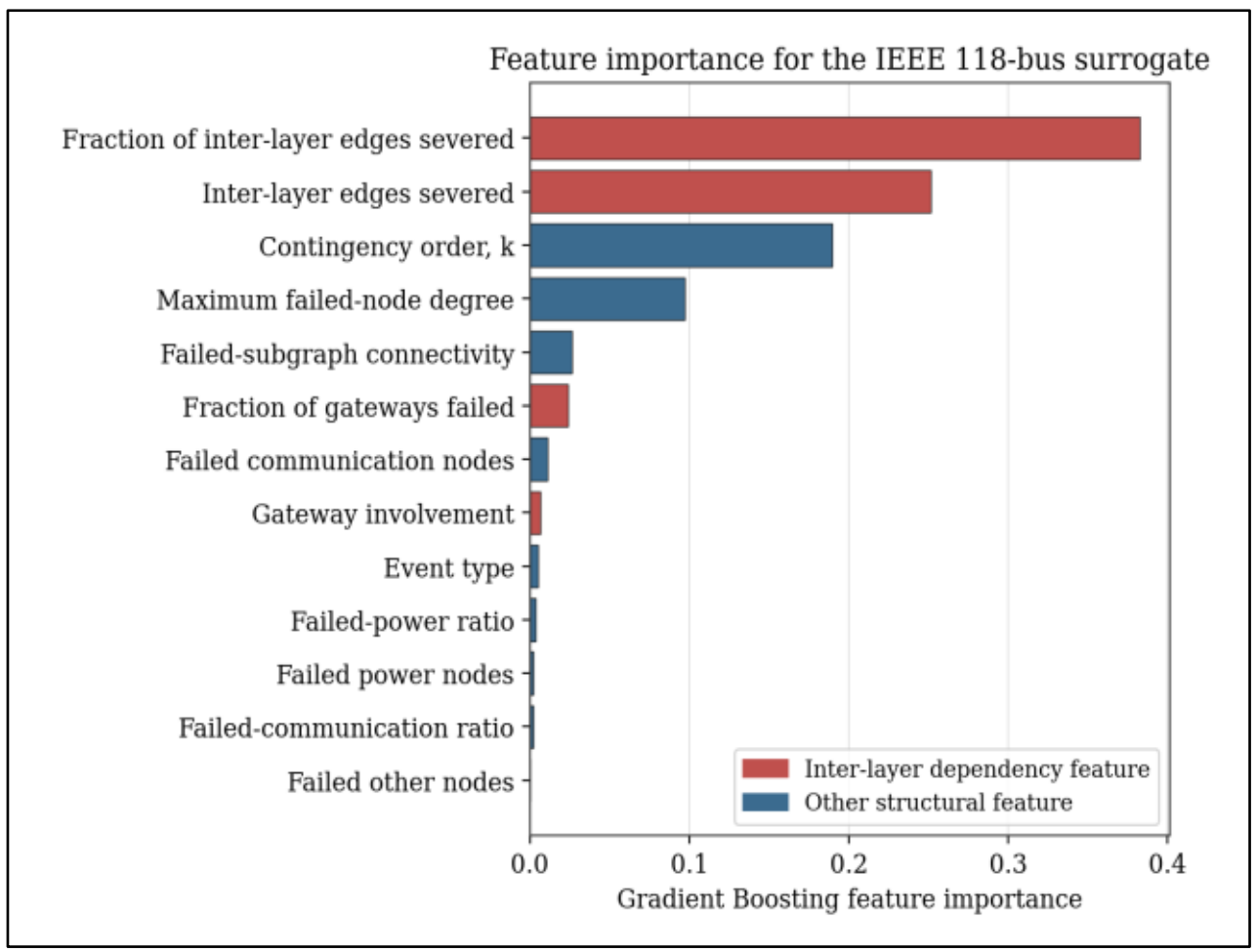

Fig. 4. Gradient Boosting feature importance on the IEEE 118-bus system. The fraction and number of severed inter-layer edges are the most influential features, indicating that the surrogate's performance is driven primarily by cross-layer disruption information rather than raw topology alone.

TABLE IV. FEATURE ABLATION ON THE IEEE 118-BUS SURROGATE.

| Feature set | # Features | Spearman | $R^2$ |
|---|---|---|---|
| Full | 13 | **0.849** | **0.689** |
| Without inter-layer features | 9 | 0.799 | 0.616 |
| Inter-layer features only | 4 | 0.763 | 0.570 |
| Contingency order $k$ only | 1 | 0.657 | 0.421 |
| Without raw-topology features | 11 | 0.813 | 0.637 |

## E. Computational Cost

TABLE V. COMPUTATIONAL COST ON THE IEEE 118-BUS DATASET

| Operation | Total time | Time per contingency |
|---|---|---|
| MIIM evaluation | 228.5 s | 21.87 ms |
| Feature extraction | 1.43 s | 0.14 ms |
| Gradient Boosting training | 0.64 s | One-time cost |
| Surrogate inference | 0.0122 s | 1.2 $\mu$s |
| End-to-end screening | 1.44 s | 0.14 ms |
| End-to-end speedup | – | $\approx 158\times$ |

Table V compares direct MIIM evaluation with surrogate-based screening on the IEEE 118-bus seed-42 dataset. Evaluating all 10,446 contingencies with MIIM requires 228.5 s, or 21.87 ms per contingency. End-to-end surrogate screening, including feature extraction and inference, requires 1.44 s, or 0.14 ms per contingency, providing an approximately $158\times$ speedup. Feature extraction accounts for most of the surrogate runtime, while inference requires only 1.2 $\mu$s per contingency. Gradient Boosting training takes 0.64 s as a one-time offline cost.

The $158\times$ figure represents the marginal screening speed once the surrogate has been trained. Including MIIM label generation, feature extraction, and model fitting, the total offline cost of 230.6 s is amortized after approximately 10,600 additional contingencies, corresponding to roughly 21,000 total evaluated cases, including the training set. As a simple linear projection from the measured average runtime, exhaustive evaluation of all $\binom{446}{3} = 14{,}686{,}780$ $N-3$ contingencies would require approximately 89 h, illustrating the larger screening regime in which surrogate prioritization becomes valuable. This projection assumes that the measured average MIIM runtime remains approximately constant across contingencies. Multiple independent runs reproduced the $158\times$ speedup, confirming its stability on the test machine. Measurements were obtained on a Lenovo Yoga 7 laptop with an 11th-generation Intel Core i5-1135G7 processor at 2.40 GHz, 12 GB RAM, and Windows 10.

### *F. Cross-Network Discussion and Workflow*

Gradient Boosting achieves Spearman $= 0.900$ on the IEEE 14-bus system and $0.849$ on the IEEE 118-bus system. The modest reduction on the larger network is expected because its severity range is narrower and its cascades are more diffuse, making contingencies harder to distinguish. Nevertheless, the surrogate maintains strong ranking performance across both network scales.

The results support a two-stage prioritization-and-verification workflow. First, the surrogate rapidly ranks contingencies and prioritizes components within a broad candidate pool. MIIM is then applied in priority order for explicit intervention analysis before protection decisions are made. Thus, the surrogate complements rather than replaces MIIM, combining approximately $158\times$ faster screening with detailed simulation-based verification.

The reported results remain protocol-specific. The $0.854$ reproducibility reference depends on the present contingency-sampling budget, and the $158\times$ speedup, while stable on the test machine, is hardware- and implementation-dependent. Broader validation within the MIIM framework across larger networks, different operating conditions, higher-order contingencies, and intervention scenarios remains future work.

## VI. Conclusion and Future Work

This paper presented a machine-learning surrogate for contingency screening and association-based component-criticality ranking in interdependent power–communication networks. Using the previously published MIIM as the label-generating cascade simulator, Gradient Boosting was trained on leakage-free structural features. On the IEEE 118-bus system, it achieved Spearman $= 0.849$ for held-out contingency-severity ranking and $0.838$ for component ranking using five-fold out-of-fold predictions.

MIIM-derived component rankings exhibit a mean cross-sample agreement of $0.854$ across three independently sampled datasets, which is treated as an empirical reproducibility reference rather than a theoretical bound. The surrogate closely approaches this reference without using in-sample predictions and provides stronger global agreement than the full-topology baselines. Feature ablation shows that its performance is driven primarily by inter-layer dependency information, while end-to-end screening is approximately $158\times$ faster than direct MIIM evaluation.

These findings support a two-stage workflow in which the surrogate rapidly screens contingencies and prioritizes components within a broad candidate pool, followed by selective MIIM evaluation of explicit intervention scenarios. The ranking does not directly identify the optimal components to harden, and the surrogate complements rather than replaces simulation-based verification. Future work will investigate richer dependency-aware representations for low-degree chokepoints, improved calibration and top-tier precision, stronger boosted-tree and graph-neural-network models, cross-network transfer, and validation on larger systems such as the 2383-bus Polish grid.